\documentclass{article}

\usepackage{microtype}
\usepackage{graphicx}
\usepackage{subfigure}
\usepackage{booktabs} 
\usepackage[utf8]{inputenc} 
\usepackage[T1]{fontenc}    
\usepackage{hyperref}       
\usepackage{url}            
\usepackage{amsfonts}       
\usepackage{nicefrac}       
\usepackage{amsmath}

\usepackage{empheq}
\usepackage{amssymb}

\usepackage[accepted]{icml2019}

\begin{document}
	
    \twocolumn[
    \icmltitle{On the Mathematical Understanding of ResNet with Feynman Path Integral}
	
	
	
    \icmlsetsymbol{equal}{*}
	
    \begin{icmlauthorlist}
    	\icmlauthor{Minghao Yin}{ts}
    	\icmlauthor{Xiu Li}{ts}
        \icmlauthor{Yongbing Zhang}{ts}
    	\icmlauthor{Shiqi Wang}{ci}
    \end{icmlauthorlist}
	
    \icmlaffiliation{ts}{Department of Automation, Tsinghua University, Shenzhen, China}
    \icmlaffiliation{ci}{City University of Hong Kong, Hong Kong, China}
	
	\icmlcorrespondingauthor{}{}
	
	\icmlkeywords{Machine Learning, ICML}
	
    \vskip 0.3in
    ]
	
	
	
	\printAffiliationsAndNotice{\icmlEqualContribution} 
	
	\begin{abstract}
		In this paper, we aim to understand Residual Network (ResNet) in a scientifically sound way by providing a bridge between ResNet and Feynman path integral. In particular, we prove that the effect of residual block is equivalent to partial differential equation, and the ResNet transforming process can be equivalently converted to Feynman path integral. These conclusions greatly help us mathematically understand the advantage of ResNet in addressing the gradient vanishing issue. More importantly, our analyses offer a path integral view of ResNet, and demonstrate that the output of certain network can be obtained by adding contributions of all paths. Moreover, the contribution of each path is proportional to $e^{-S}$, where $S$ is the action given by time integral of Lagrangian $L$. This lays the solid foundation in the understanding of ResNet, and provides insights in the future design of convolutional neural network architecture. Based on these results, we have designed the network using partial differential operators, which further validates our theoritical analyses.
	\end{abstract}
	
	\section{Introduction}
	Recently, dramatic progress has been made in the design of neural network architecture in deep learning, and convolutional neural network (CNN) \cite{krizhevsky2012imagenet} \cite{simonyan2014very} plays a prominent role in many tasks due to its effectiveness and practicability. To further improve the performance of CNN, the residual network (ResNet) \cite{he2016deep} was proposed in 2016, which adds the skip connection between different layers. The structure of convolution kernels accompanied with the skip connection is called  residual block, with the help of which vanishing gradient can be successfully avoided. In this manner, the CNN can be greatly deepened based on a well-designed network architecture to achieve promising performance. Despite its great success, the mathematical understanding of ResNet is still lacking, which plays important roles in the design of future network architecture.
	
\begin{figure}[]
		\centering
		\subfigure[]{
			\includegraphics[width=2in]{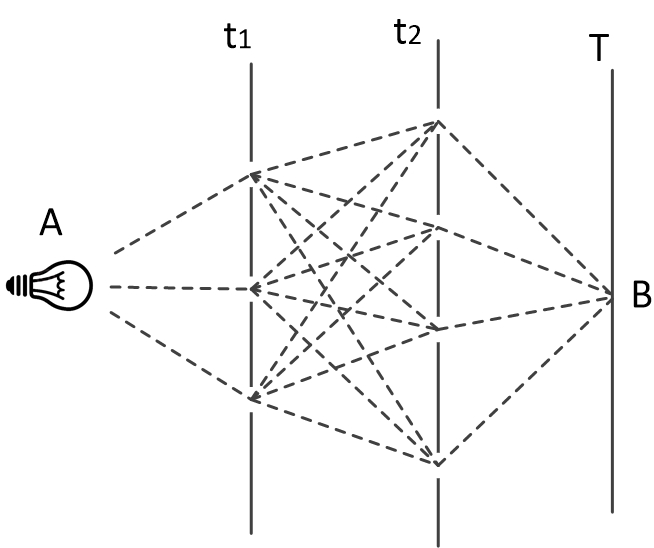}}
		\hspace{0.2in}
		\subfigure[]{
			\includegraphics[width=2.3in]{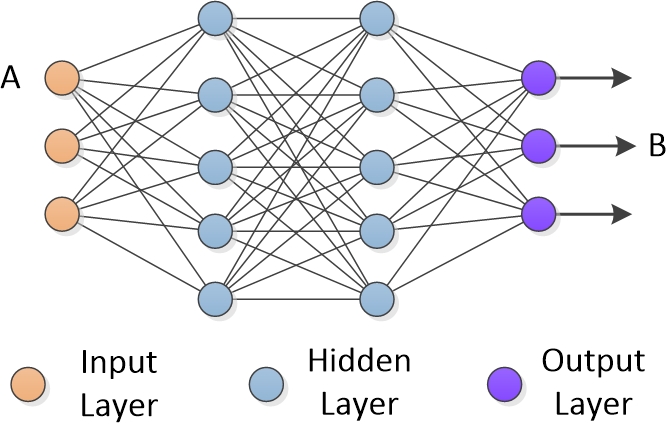}}
		\caption{(a) The possible trajectories of photon passing through two baffles with slits. (b) The neural network with two hidden layers.}
		\label{fig:path_and_DNN}
	\end{figure}

	The Feynman path integral is a formulation of quantum mechanics describing the evolution process of quantum system  \cite{dirac1933lagrangian} \cite{van1928correspondence}. In principle, the formulation of path integral is equivalent to the formulation of quantum physics (Schrödinger equation) \cite{feynman2005space} \cite{feynman2010quantum}. For instance, we can consider the moving quantum particle whose movement is controlled by Schrödinger equation as an example. The particle has no certain position or momentum as it acts as a random variable with probability distribution in space. When it moves over time, the particle has various possible trajectories as potential candidates. As such, all possible trajectories should be taken into consideration, and every path contributes its own to the final state. Therefore, the contributions from all historical paths have to be added up to obtain the final state. We demonstrate this phenomenon with the light emitting photons. As shown in Fig.~\ref{fig:path_and_DNN}(a), every photon fired by the light has to pass two layers of baffle with slits. Behind the two baffles there is a screen $T$ which receives the photons. Baffle $t_{1}$ has three slits, such that the photon has three possible choices. Moreover, baffle $t_{2}$ has four slits, such that from baffle $t_{1}$ to baffle $t_{2}$ the photon has $3 \times 4=12$ possible paths. To predict how many photons will travel to the point $B$, every possible path has to be involved. 
	
\begin{figure}
		\centering
		\includegraphics[width=3.0in]{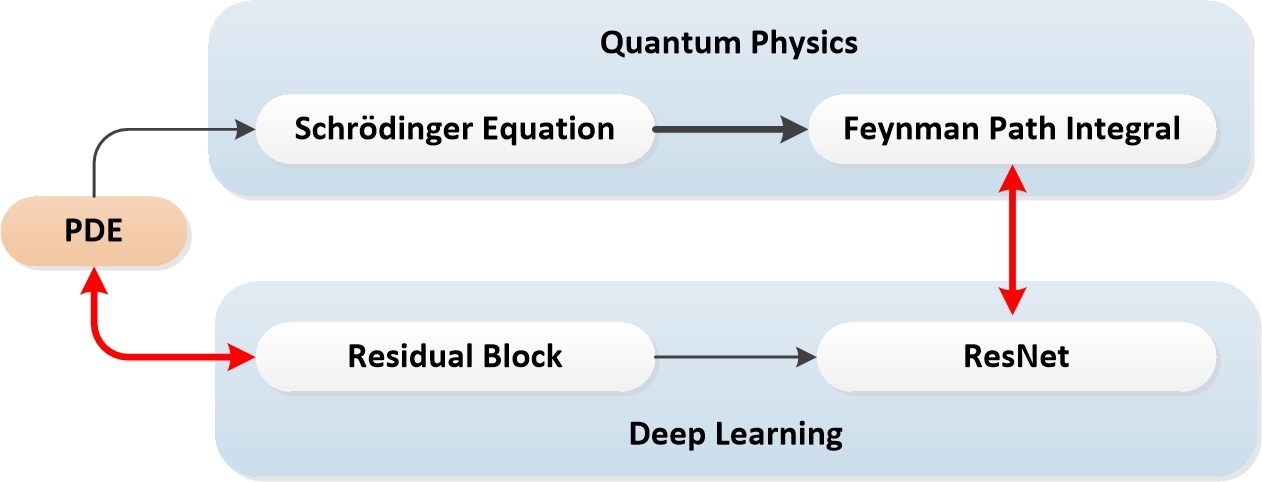}
		\caption{Illustration of the relationship between deep neural network and Feynman path integral.}
		\label{fig:architecture}
	\end{figure}

	As shown in Fig.~\ref{fig:path_and_DNN}, in analogy to path integral, 
	every neuron in DNN receives the inputs from all the previously connected neurons. This indicates an implicit relationship between deep learning and Feynman path integral. In view of this, we further prove the ResNet is equivalent to Feynman path integral, such that the design of ResNet can be interpreted from a new perspective. In Fig.~\ref{fig:architecture}, the relationship between the quantum physics and deep learning is established based on the partial differential equation (PDE) with the following three arguments,
	\begin{itemize}
		\item The evolution of solution to Schrödinger equation can be transformed to Feynman path integral formulation \cite{feynman1948rp}.
		\item A residual block with skip connection is equivalent to a partial differential equation.
		\item The ResNet can be converted to the formulation of path integral.
		
	\end{itemize}
	In this paper, we show that the study of quantum physics leads to the mathematical explanations of the interesting properties of ResNet regarding the vanishing gradient problem. Based on theoretical analyses, we explore the advantages of ResNet experimentally and establish the network with partial differential operators to validate our derivations. We assert that such explanations are of fundamental importance to the future design and applications of CNN models.
	
	\section{Background}
	\subsection{Schrödinger Equation}
	For a system in quantum physics, the Schrödinger equation \cite{schrodinger1926undulatory} describes how it changes over time, and the most general time-dependent form of Schrödinger equation which describes the evolution of a quantum particle can be written as follows,
	\begin{equation}
	i \frac{\partial}{\partial t} u(x,t) = \hat{H} u(x,t)=\left(\frac{\hat{p}^{2}}{2m}+V(x,t)\right) u(x,t)
	\label{con:schrodinger}
	\end{equation}
	Here, we set plank constant $\hbar=1$ for convenience. The notation $u(x,t)$ is the wave function of quantum particle, $\hat{H}$ is called Hamiltonian operator which characterizes the total energy of the particle (kinetic energy plus potential energy), $\frac{\hat{p}^{2}}{2m}$ is the kinetic energy operator, and $V(x,t)$ is the potential energy. The momentum operator $\hat{p}$ denotes $-i \frac{\partial}{\partial x}$ in position space. When performing Fourier transform on the momentum operator, the real momentum in frequency space can be obtained, which is also called momentum space in physics,
	\begin{equation}
	\begin{aligned}
	\hat{p} u(x,t)=-i \frac{\partial}{\partial x} u(x,t) &\Longleftrightarrow p \tilde{u}(p,t)  \\
	\frac{\hat{p}^{2}}{2m} u(x,t) =\frac{(-i \frac{\partial}{\partial x})^{2}}{2m} u(x,t)  &\Longleftrightarrow \frac{p^{2}}{2m}  \tilde{u}(p,t) 
	\end{aligned}
	\end{equation}
	where $\tilde{u}(p,t)$ stands for the Fourier transform of function $u(x,t)$ in momentum space. After Fourier transform, $T=\frac{p^{2}}{2m}$ is exactly the kinetic energy of the quantum particle, and $\hat{T}=\frac{\hat{p}^{2}}{2m}$ is also regarded as the kinetic energy operator in position space. 
	
	Generally speaking, representations in position space (time domain) and momentum space (frequency domain) are interchangeable, and different set of vector bases in Hilbert space lead to different representations. Based on the Dirac bra-ket \cite{dirac1939new} notation for describing quantum states, the vector base of position space is denoted as $|x\rangle$, and the vector base of momentum space is denoted as $|p \rangle$. For any state $|\phi \rangle$, its conjugate transpose is written as $\langle \phi|$, $\langle \phi|\phi \rangle=|\phi|^{2}$, and its representations in position and momentum space are $|\phi_{x} \rangle = \langle x | \phi \rangle$ and $|\phi_{p} \rangle = \langle p | \phi \rangle$, respectively. It is also worth mentioning that in quantum system, for states or operators there is no need to specify which set of vector bases are used. Position and momentum serve as different representations of the same process and states. As such, the following conclusions can be obtained,
	\begin{equation}
	\begin{aligned}
	u(x,t)=\langle x|u(t)\rangle &\qquad \tilde{u}(p,t)=\langle p|u(t) \rangle \\
	\langle x'|x\rangle=\delta(x'-x) &\qquad \langle p'|p \rangle=\delta(p'-p) \\
	\int_{x} \mathrm{d}x |x\rangle \langle x|=1 &\qquad \int_{p} \mathrm{d}p |p\rangle \langle p| =1
	\end{aligned}
	\end{equation}
	and the inner product of $|x \rangle $ and $|p \rangle$ naturally defines the Fourier transform:
	\begin{equation}
	\begin{aligned}
	&\langle x|p \rangle =\frac{1}{2\pi} e^{ipx} \qquad \langle p|x \rangle = e^{-ipx} \\
	&\tilde{u}(p,t) = \langle p|u(t) \rangle =\int_{x} \mathrm{d}x \langle p|x \rangle \langle x|u(t) \rangle = \int_{x} \mathrm{d}x e^{-ipx} u(x,t)
	\end{aligned}
	\end{equation}
	Without specific specifications of the representation bases, the Schrödinger equation and its solution are:
	\begin{equation}
	i \frac{\partial}{\partial t} |u(t) \rangle =H |u(t) \rangle \qquad | u(t) \rangle =e^{-itH} | u(0 )\rangle
	\label{con:Sch_2}
	\end{equation}
	\begin{figure}
		\centering
		\subfigure[]{
			\includegraphics[width=1.5in]{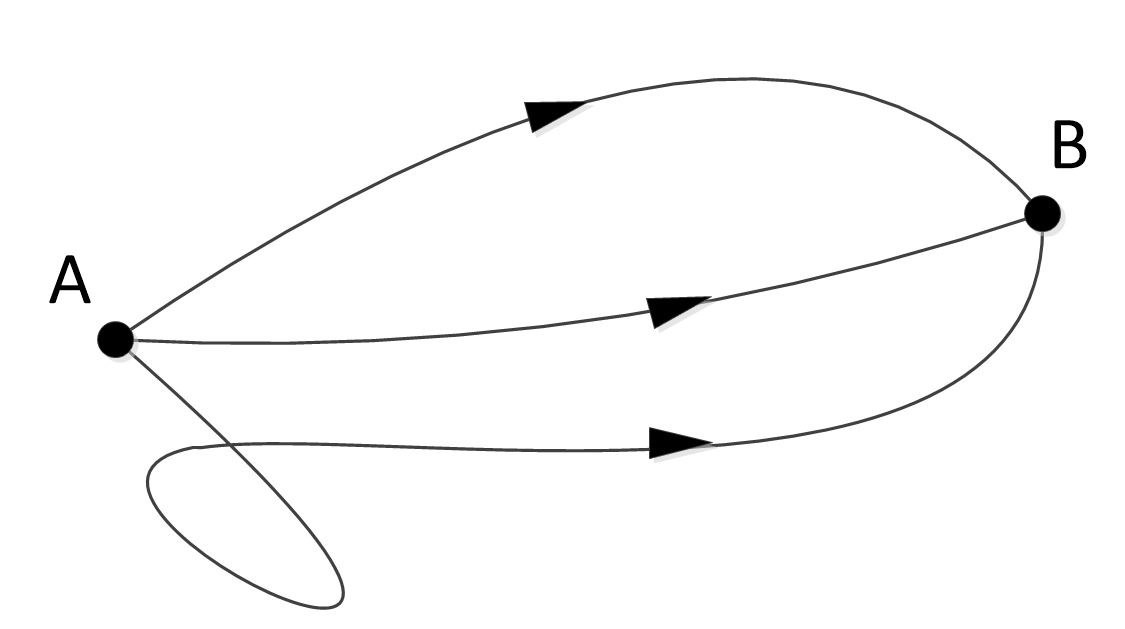}}
		\subfigure[]{
			\includegraphics[width=2in]{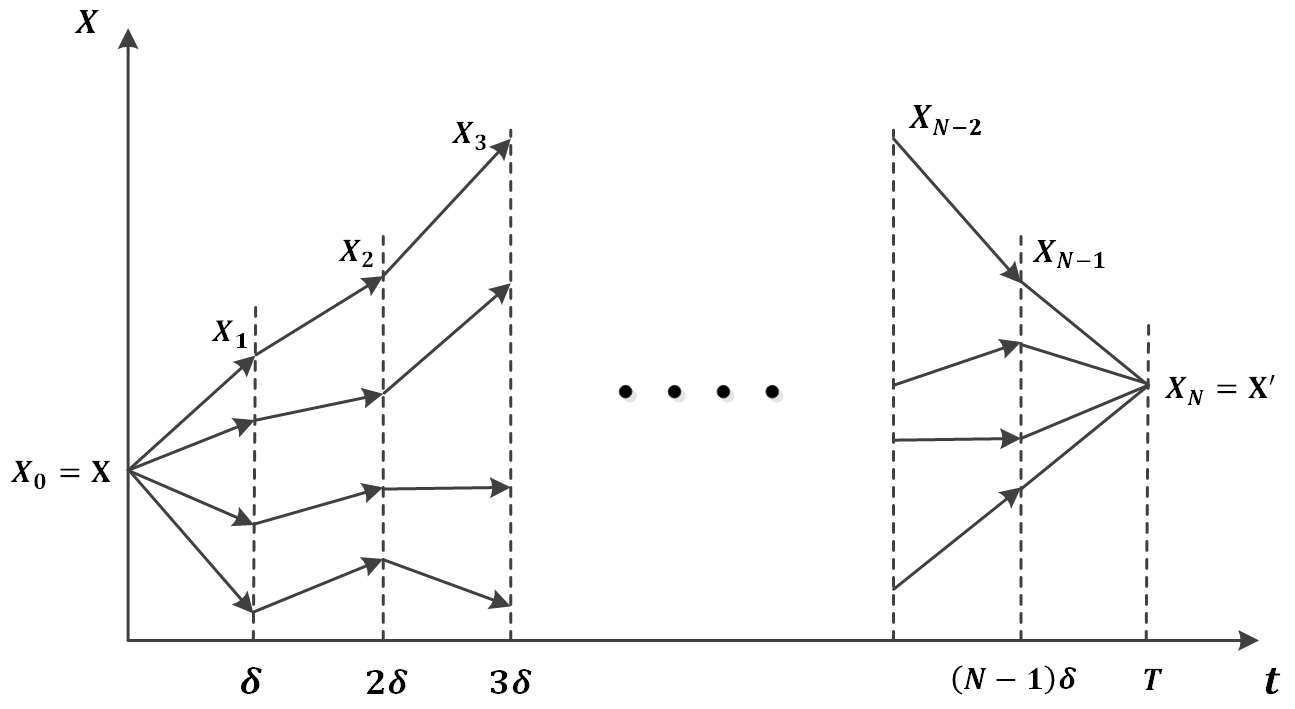}}
		\caption{(a) Candidate paths for the particle from position A to B. (b) Step by step particle movement prediction by cutting long time period into pieces. }
		\label{fig:3path}
	\end{figure}
	\subsection{Feynman Path Integral}
	Feynman path integral, which was developed in 1948 by Richard Feynman \cite{feynman1948rp}, provides us a new perspective regarding the mechanics of a single quantum particle. Generally speaking, the path integral formulation of a quantum particle is equal to its Schrödinger equation \cite{feynman2005space} \cite{mackenzie2000path}.
	In particular, a particle in quantum physics has no certain position or momentum, and can only be characterized with probability distribution in space. Moreover, when the particle moves in space, it is impossible to clarify its specific trajectory. As shown in Fig.~\ref{fig:3path}, from position A to position B, the particle has many possible paths. Unlike the classical mechanics, every possible path should be taken into consideration. Feynman path integral aims to integrate effects from all possible paths derived from the Schrödinger equation. From Eqn.(\ref{con:Sch_2}), we have,
	\begin{equation}
	\begin{aligned}
	u(x',T)&=\langle x'|u(T) \rangle=\langle x' | e^{-iTH}|u(0) \rangle \\
	&=\int_{x} \mathrm{d}x \langle x' |e^{-iTH} |x \rangle \langle x|u(0) \rangle \\
	&=\int_{x} \mathrm{d}x \langle x' |e^{-iTH} |x \rangle u(x,0)
	\end{aligned}
	\end{equation}
	Here, $K=\langle x' |e^{-iTH} |x \rangle$ is called propagator between $x$ and $x'$. With the propagator, we can calculate the evolution process of the particle state. However, propagator $K$ is a complex function and is hard to calculate, such that the time $T$ can be divided into $N$ pieces ($\delta=T/N$) as follows,
	\begin{small}
		\begin{equation}
		\begin{aligned}
		K&=\langle x' | (e^{-i\delta H})^{N} |x \rangle \\
		&=\int \mathrm{d}x_{1} \cdots \mathrm{d}x_{N-1} \langle x' |e^{-i\delta H} |x_{N-1}\rangle \\ 
		& \quad \langle x_{N-1} |e^{-i\delta H} |x_{N-2}\rangle \cdots \langle x_{1} |e^{-i\delta H} |x\rangle \\
		&=\int \mathrm{d} x_{1} \cdots \mathrm{d}x_{N-1} K_{x',x_{N-1}} K_{x_{N-1},x_{N-2}} \cdots K_{x_{1},x}
		\label{con:path_integral1}
		\end{aligned}
		\end{equation}
	\end{small}
	As shown in Fig.~\ref{fig:3path}(b), the calculation of $K$ computes the integrals through all paths that start from $x$ and end at $x'$. First of all,  let us consider the propagator between a small time interval using Taylor-series expansion based on the assumption that $\delta$ is very small,
	\begin{equation}
	\begin{aligned}
	K_{x_{j+1},x_{j}}&= \langle x_{j+1} | e^{-i\delta H}| x_{j} \rangle 
	\approx \langle x_{j+1} | 1-i\delta H | x_{j} \rangle \\
	&=\langle x_{j+1}|x_{j}\rangle -i \delta \langle x_{j+1} | H | x_{j} \rangle
	\end{aligned}
	\end{equation}
	Based on Fourier transform, we have $\langle x_{j+1}|x_{j} \rangle$ in momentum space:
	\begin{equation}
	\langle x_{j+1}|x_{j} \rangle=\delta(x_{j+1}-x_{j})=\frac{1}{2 \pi}\int \mathrm{d}p_{j} e^{ip_{j}(x_{j+1}-x_{j})}
	\label{con:delta_f}
	\end{equation}
	For the second part of  $K_{x_{j+1},x_{j}}$, again we can expand it on momentum space:
	\begin{small}
		\begin{equation}
		\begin{aligned}
		\langle x_{j+1} | H | x_{j} \rangle &= \int \mathrm{d} p_{j} \langle x_{j+1}|p_{j}\rangle \langle p_{j}| H |x_{j} \rangle \\
		&=\int \mathrm{d} p_{j} \langle x_{j+1}|p_{j}\rangle \langle p_{j}| x_{j} \rangle H(p_{j},x_{j}) \\
		&=\int \frac{\mathrm{d} p_{j}}{2 \pi} e^{ip_{j}x_{j+1}} e^{-ip_{j}x_{j}} H(p_{j},x_{j}) \\
		&=\int \frac{\mathrm{d} p_{j}}{2 \pi} e^{ip_{j}(x_{j+1}-x_{j})} H(p_{j},x_{j}) 
		\label{con:hamilton_f}
		\end{aligned}
		\end{equation}
	\end{small}
By combining Eqn.~(\ref{con:delta_f}) and Eqn.~(\ref{con:hamilton_f}) together, we can obtain the propagator $K_{x_{j+1},x_{j}}$ in a short time interval $\delta$:
	\begin{equation}
	\begin{aligned}
	K_{x_{j+1},x_{j}}&=\int \frac{\mathrm{d}p_{j}}{2 \pi} e^{ip_{j}(x_{j+1}-x_{j})}(1-i\delta H(p_{j},x_{j})) \\
	&=\int \frac{\mathrm{d} p_{j}}{2 \pi} e^{ip_{j}(x_{j+1}-x_{j})-i\delta H(p_{j},x_{j}) }
	\end{aligned}
	\end{equation}
	Assuming the particle speed is $\dot{x}_{j}=(x_{j+1}-x_{j})/\delta$, we can have,
	\begin{equation}
	K_{x_{j+1},x_{j}}=\int \frac{\mathrm{d} p_{j}}{2 \pi} e^{i\delta(p_{j}\dot{x}_{j}-H(p_{j},x_{j}))}
	\label{con:small_propagator}
	\end{equation}
	By incorporating $K_{x_{j+1},x_{j}}$ from Eqn.~(\ref{con:small_propagator}) into Eqn.~(\ref{con:path_integral1}), the propagator during a long time interval $T$ is formulated as,
	\begin{small}
		\begin{equation}
		\begin{aligned}
		K&=\int \prod_{i} \mathrm{d}x_{i}  \int \frac{\mathrm{d}p_{1}}{2 \pi} \cdots \frac{\mathrm{d} p_{N-1}}{2 \pi} \prod_{j=0}^{N-1} e^{i\delta(p_{j}\dot{x}_{j}-H(p_{j},x_{j}))} \\
		&=\int \prod_{i} \mathrm{d}x_{i} \int \frac{\mathrm{d}p_{1}}{2 \pi} \cdots \frac{\mathrm{d} p_{N-1}}{2 \pi} e^{i \sum_{j=0}^{N-1}\delta(p_{j}\dot{x}_{j}-H(p_{j},x_{j}))}
		\label{con:all_propagator}
		\end{aligned}
		\end{equation}
	\end{small}
	When number of time pieces $N \to \infty$, the propagator $K$ becomes:
	\begin{equation}
	K=\int Dx(t) Dp(t) e^{i\int \mathrm{d}t (p\dot{x}-H(p,x))}
	\label{con:phase_space_integral}
	\end{equation}
	The propagator $K$ computes the integrals over trajectory functions $x(t)$ and $p(t)$. It is a functional integral, and function $x(t)$ satisfies $x(0)=x$, $x(T)=x'$. Eqn.~(\ref{con:phase_space_integral}) is called the phase-space path integral, and the propagator computes the integral paths not only in position space but also in momentum space.
	
	For quantum particle movement, the hamiltonian $H(p,x)= \frac{p^2}{2m} + V(x)$, and it is integrable in momentum space. Moreover, by incorporating $H(p,x)$ into Eqn.~(\ref{con:all_propagator}), we will have,
	\begin{equation}
	\begin{aligned}
	\int \frac{\mathrm{d}p}{2 \pi} e^{i\delta}(p\dot{x}-p^{2}/2m)=\sqrt{\frac{m}{2\pi i \delta}} e^{i \delta m \dot{x}^{2}/2}
	\end{aligned}
	\end{equation}
	\begin{small}
		\begin{equation}
		\begin{aligned}
		K&=\int \mathrm{d}x_{1} \cdots \mathrm{d}x_{N-1} e^{-i\delta \sum_{j=1}^{N-1} V(x_{j})} \\
		&\quad \int \frac{\mathrm{d}p_{0}}{2 \pi} \cdots \frac{\mathrm{d}p_{N-1}}{2 \pi} e^{i\delta \sum_{j=0}^{N-1}(p_{j}\dot{x}_{j}-p_{j}^{2}/2m)} \\
		&=(\frac{m}{2 \pi i \delta})^{\frac{N}{2}} \int \mathrm{d}x_{1} \cdots \mathrm{d}x_{N-1} e^{i\delta \sum_{j=0}^{N-1}(m \dot{x}_{j}^{2}/2-V(x_{j})}
		\end{aligned}
		\end{equation}
	\end{small}
	It is interesting to see that $m \dot{x}^{2}/2 -V(x)$ is the kinetic energy $T=m \dot{x}^{2}/2$ minus potential energy $V(x)$ of the system. As a matter of fact, it is exactly the Lagrangian $L=T-V$ of the system. Theoretically speaking, the Lagrangian $L$ also equals to the Legendre transform of Hamiltonian: $p\dot{x}-H(p,x)$, which aligns with Eqn.~(\ref{con:phase_space_integral}). As such, the appearance of Lagrangian is inevitable. Finally, the ``action" $S$ over a specific path function $x(t)$ is defined as the time integral of Lagrangian $L(x,\dot{x},t)$,
	\begin{equation}
	\begin{aligned}
	K=\int Dx(t)e^{i\int \mathrm{d}t L(x,\dot{x},t)}=\int Dx(t)e^{i S(x(t))}
	\label{con:quantum_path_integral}
	\end{aligned}
	\end{equation}
	From the above Feynman path integral formulation, we can have the following conclusions,
	\begin{itemize}
		\item The final state can be obtained by adding the contributions of all paths in the configuration space together.
		\item The contribution of a path is proportional to $e^{iS}$, where $S$ is the action given by the time integral of the Lagrangian $L$ along the path.
	\end{itemize}
	\section{From PDE to Residual Block}
	\label{headings}
	
	In this section, we establish the relationship between PDE and convolutional residual block, and demonstrate that they are mathematically equivalent to each other. 
	This helps us understand why ResNets could be deeper in order to achieve promising performance in various tasks.
	First of all, let us consider a two order PDE:
	\begin{equation}
	\frac{\partial u(x,t)}{\partial t} = \frac{1}{2} \sigma^{2} \frac{\partial^{2} u(x,t)}{\partial x^{2}} + b \frac{\partial u(x,t)}{\partial x} + c u(x,t)
	\label{con:PDE}
	\end{equation}
	and the discrete form of Eqn.~(\ref{con:PDE}) can be written as,
	\begin{small}
		\begin{equation}
		\begin{split}
		&u(x,t+1)-u(x,t)= \\
		&\frac{1}{2} \sigma^{2} (u(x+1,t)-2u(x,t)+u(x-1,t)) \\
		&+ \frac{b}{2} (u(x+1,t)-u(x-1,t)) + cu(x,t)
		\end{split}
		\label{con:DPDE}
		\end{equation}
	\end{small}
	It is convenient to rewrite Eqn.~(\ref{con:DPDE}) into convolution form:
	\begin{small}
		\begin{equation}
		\begin{aligned}
		&u(x,t+1)= \\
		&u(x,t)+[\frac{1}{2}(\sigma^{2}+b),c-\sigma^{2},\frac{1}{2}(\sigma^{2}-b)] \ast u(x,t) \\
		\label{con:DPDE2}
		\end{aligned}
		\end{equation}
	\end{small}
	By regarding the convolution kernel $[\frac{1}{2}(\sigma^{2}+b),c-\sigma^{2},\frac{1}{2}(\sigma^{2}-b)]$ as $w(x,t)$, Eqn.~(\ref{con:DPDE2}) has the identical form as a residual block,
	\begin{equation}
	u(x,t+1)=u(x,t)+w(x,t) \ast u(x,t)
	\label{con:iter1}
	\end{equation}
	From the above analysis it is found that any two order PDE can be rewritten as a residual block with the corresponding convolution kernel size equaling to 3, and higher order PDE can be converted to the residual block with larger convolution kernel. As the kernel size of ResNet is usually small, there exists a correspondence between the residual block and the two order PDE. Moreover, Fourier transform is widely used in solving PDE, and it can also clarify the reason why deeper ResNet leads to better performance as previously stated. By applying Fourier transform to Eqn.~(\ref{con:PDE}) and replacing every operator $\frac{\partial}{\partial x}$ by $ i \lambda$, we have, 
	\begin{equation}
	\hat{T}_{x}=\frac{1}{2}\sigma^{2}\frac{\partial^{2}}{\partial x^{2}}+ b\frac{\partial}{\partial x} +c \\
	\Longleftrightarrow \hat{T}_{p}=-\frac{1}{2} \sigma^{2} p^{2} + ib p + c
	\end{equation}
	\begin{equation}
	\hat{T}_{p} \tilde{u}(p,t) =\frac{\mathrm{d}}{\mathrm{dt}} \tilde{u}(p,t) 
	\label{con:FPDE}
	\end{equation}
	The solution of Eqn.~(\ref{con:FPDE}) is,
	\begin{equation}
	\begin{aligned}
	\tilde{u}(p,t)=e^{\hat{T}_{p}t}\tilde{u}(p,0)
	\end{aligned}
	\label{con:FPDES}
	\end{equation}
	Assuming time $t$ is small enough, we can obtain,
	\begin{equation}
	\tilde{u}(p,t)\approx(1+\hat{T}_{p}t)\tilde{u}(p,0)
	\label{con:FPDES2}
	\end{equation}
	By applying inverse Fourier transform to Eqn.~(\ref{con:FPDES2}), from convolution theorem we can have the following relationship,
	\begin{equation}
	\begin{aligned}
	u(x,t)=u(x,0)+t\hat{T}_{x}\delta(x) \ast u(x,0)
	\label{con:iter2}
	\end{aligned}
	\end{equation}
	It is obvious that Eqn.~(\ref{con:iter1}) and Eqn.~(\ref{con:iter2}) are equivalent, and the convolution kernel $w(x,t)$ corresponds to $t\hat{T}_{x}\delta(x)$ in continuous space. Basically, the relationship between PDE and ResNet is characterized by Eqn.~(\ref{con:iter1}) from a discrete perspective of view, and Eqn.~(\ref{con:iter2}) grasps the essence of this relationship in the  continuous domain. It is also widely acknowledged that the numerical values of the convolution kernels in residual blocks are extremely small. This aligns with the assumption that the time $t$ is small, as small time period $t$ ensures that the corresponding convolution kernel $t\hat{T}_{x}\delta(x)$ is numerically small from Eqn.~(\ref{con:FPDES}) to Eqn.~(\ref{con:FPDES2}). Moreover, the small size convolutional kernels results in low order PDE, which simplifies evolution of the iterative system. The investigation of the relationship between PDE and residual block also sheds light on better understanding of the ResNet.
	
	
	
	
	
	
	
	
	\section{On the Understanding of ResNet with Path-Integral Formula}
	\label{others}
	In this section, we aim to better understand the philosophy behind ResNet with Feynman Path-Integral. For a certain ResNet, feature maps at layer $t$ is represented as $u_{t}$ . First of all, update process between $u_{t-1}$ and $u_{t}$ is considered. In a residual block, assuming $f$ is the skip connection weight and $w$ is the convolution kernel, $u_{t-1}$ can be transformed into $u_{t}$ as follows,
	\begin{equation}
	u(x_{t})=relu  \{ f \cdot u(x_{t-1})+w(x) \ast u(x_{t-1})\}
	\label{con:res_path1}
	\end{equation}
	Given $f \cdot u+w \ast u = f(\delta + w/f) \ast u$, Eqn.~(\ref{con:res_path1}) can be rewritten as follows,
	\begin{equation}
	u(x_{t})= \sum_{x_{t-1}} \kappa \circ f(\delta(x_{t}-x_{t-1})+\Omega(x_{t}-x_{t-1}))u(x_{t-1})
	\label{con:res_path2}
	\end{equation}
	In Eqn.~(\ref{con:res_path2}), $\Omega(x)=\frac{w(x)}{f}$. Under the influence of $relu$, $\kappa$ deployed on $f$ leads to $|f|$ or $0$ depending on $u(x)$. As such, we can define $h_{t}=\log(\kappa \circ f)$. In analogies to Eqn.~(\ref{con:delta_f}) and Eqn.~(\ref{con:hamilton_f}), with inverse discrete Fourier transform we can obtain functions in the frequency domain as follows,
	\begin{equation}
	\begin{aligned}
	\delta(x_{t}-x_{t-1})&=\frac{1}{M}\sum_{k=0}^{M-1}e^{i\frac{2\pi}{M}k(x_{t}-x_{t-1})}\\
	&=\sum_{p_{t-1}}e^{ip_{t-1}(x_{t}-x_{t-1})}
	\label{con:deltaF}
	\end{aligned}
	\end{equation}
	\begin{equation}
	\begin{aligned}
	\Omega(x_{t}-x_{t-1})&=\frac{1}{M}\sum_{k=0}^{M-1}e^{i\frac{2\pi}{M}k(x_{t}-x_{t-1}))}{\Omega}(k) \\
	&=\sum_{p_{t-1}}e^{ip_{t-1}(x_{t}-x_{t-1})}\tilde{\Omega}(p_{t-1})
	\label{con:omegaF}
	\end{aligned}
	\end{equation}
	By incorporating Eqn.~(\ref{con:deltaF}) and Eqn.~(\ref{con:omegaF}) into Eqn.~(\ref{con:res_path2}), we have,
	\begin{small}
		\begin{equation}
		\begin{aligned}
		u(x_{t})=\sum_{x_{t-1}} \sum_{p_{t-1}}e^{ip_{t-1}(x_{t}-x_{t-1})}e^{h_{t}}(1+\tilde{\Omega}(p_{t-1}))u(x_{t-1})
		\end{aligned}
		\end{equation}
	\end{small}
	Based on the assumption that the skip connection weights are much larger than the convolution kernel, we have $ f_{t} \gg w(x_{t})$, such that $\Omega(x_{t}) \ll 1 $ and $1+\Omega(x_{t}) \approx e^{\Omega(x_{t})}$, leading to,
	\begin{equation}
	\begin{aligned}
	u(x_{t})= \sum_{x_{t-1}} \sum_{p_{t-1}}e^{ip_{t-1}(x_{t}-x_{t-1})+h_{t}+\tilde{\Omega}(p_{t-1})}u(x_{t-1})
	\label{con:temp1}
	\end{aligned}
	\end{equation}
	With the definition of Hamiltonian $H(p_{t-1})=-\tilde{\Omega}(p_{t-1}) - h_{t}$, Eqn.~(\ref{con:temp1}) can be rewritten as:
	\begin{equation}
	u(x_{t})= \sum_{x_{t-1}} \sum_{p_{t-1}}e^{ip_{t-1}(x_{t}-x_{t-1}) - H(p_{t-1})}u(x_{t-1})
	\end{equation}
	This corresponds to Eqn.~(\ref{con:small_propagator}), where the small time propagator $K_{x_{j+1},x_{j}}$ is defined. In essence, $H$ defined energy of a certain path. After $N$ residual convolution steps, the outputs from the network can be formulated as,
	\begin{equation}
	\begin{aligned}
	u(x_{N})&=\prod_{t=1}^{t=N} \sum_{x_{t-1}} \sum_{p_{t-1}}e^{ip_{t-1}(x_{t}-x_{t-1}) - H(p_{t-1})}u(x_{0})  \\
	&= \sum_{x_{path}} \sum_{p_{path}}\prod_{t}e^{ip_{t-1}(x_{t}-x_{t-1}) - H(p_{t-1})}u^{\phi_{0}}(x_{0}) 
	\label{con:res_path3}
	\end{aligned}
	\end{equation}
	In Eqn.~(\ref{con:res_path3}), we define $\prod_{t} \sum_{x_{t-1}} = \sum_{x_{path}} \prod_{t}$, where $\sum_{x_{path}}=\sum_{x_{0}}\sum_{x_{1}}...\sum_{x_{N-1}}$. This implies that summing along every path leads to $x_{N}$. More strictly speaking, $\sum_{x_{path}}$ is the functional integral over trajectory functions. Analogously, $\sum_{p_{path}}$ can be derived. Eqn.~(\ref{con:res_path3}) can be equivalently written as follows, 
	\begin{equation}
	u(x_{N})= \sum_{x_{path}} \sum_{p_{path}}e^{\sum_{t}(ip_{t-1}(x_{t}-x_{t-1})-H(p_{t-1}))}u(x_{0})
	\label{con:res_path4}
	\end{equation}
	By comparing Eqn.~(\ref{con:res_path4}) to Eqn.~(\ref{con:phase_space_integral}), obviously Eqn.~(\ref{con:res_path4}) can be regarded as the phase space path integral of ResNet. Regarding the mathematical equivalence between residual convolution and PDE as discussed in Eqn.~(\ref{con:FPDES}), the residual convolution results in $e^{\hat{T}_{p}t}$ in the frequency domain. Obviously, $\hat{T}_{p}=-\frac{1}{2} \sigma^{2} p^{2} + ib p + c$ corresponds to $H(p_{t-1})$:
	\begin{equation}
	H(p_{t-1})=\frac{1}{2} \sigma^{2} p^{2}_{t-1} - ib p_{t-1} - c
	\end{equation}
	The two order form of Hamiltonian $H$ guarantees that it is integrable by inverse Fourier transform over frequency $p$, such that a path integral formula in position space can be obtained,
	\begin{equation}
	\begin{aligned}
	u(x_{N})&= \sum_{x_{path}} \sum_{p_{path}}\prod_{t}e^{ip_{t-1}(x_{t}-x_{t-1})-H(p_{t-1})}u(x_{0})\\
	&=\sum_{x_{path}}\prod_{t}\sum_{p_{t-1}}e^{ip_{t-1}(x_{t}-x_{t-1})-H(p_{t-1})}u(x_{0}) \\
	&=\sum_{x_{path}}\prod_{t} \frac{e^{c}}{\sigma} e^{-{(x_{t}-x_{t-1}+b)^{2}}/{2\sigma^{2}}}u(x_{0})
	\end{aligned}
	\end{equation}
	
	By defining $\dot{x}=x_{t}-x_{t-1}$, with the definition of kinetic energy $T$ and potential energy $V$, the Lagrangian $L=T-V$ can be obtained,
	\begin{equation}
	V=c-log(\sigma)
	\label{con:potiential_energy}
	\end{equation}
	\begin{equation}
	T(\dot{x})=(\dot{x}+b)^{2}/2\sigma^{2}
	\label{con:kinect_energy}
	\end{equation}
	As such, the evolution of ResNet can be written based on the form of integrals over action $S$:
	\begin{equation}
	\begin{aligned}
	u(x_{N})&= \sum_{x_{path}} e^{\sum_{t} V-T(\dot{x})}u(x_{0}) \\
	&= \sum_{x_{path}}e^{-\sum_{t} L^{t}_{path}}u(x_{0}) \\
	&= \sum_{x_{path}}e^{-S_{path}}u(x_{0})
	\label{con:convolution_path_integral}
	\end{aligned}
	\end{equation}
	This formulation is equivalent to the Feynman path integral formulation in Eqn.~(\ref{con:quantum_path_integral}), such that it is regarded as the path integral formulation of ResNet which helps us better understand the ResNet:
	\begin{itemize}
		\item The output of ResNet is given by adding the contributions along all paths that information flow through together.
		\item The contribution of a path is proportional to $e^{-S_{path}}$, where $S_{path}$ is the action given by the time integral of the Lagrangian $L^{t}_{path}$ along the path. Lagrangian $L^{t}_{path}$ is defined based on the kinetic energy $T^{t}_{path}$(Eqn.~(\ref{con:kinect_energy})) and potential energy $V^{t}_{path}$(Eqn.~(\ref{con:potiential_energy})), i.e., $L^{t}_{path}=T^{t}_{path}-V^{t}_{path}$.
	\end{itemize}
	\begin{figure}[]
		\centering
		\subfigure[]{
			\includegraphics[width=0.8in]{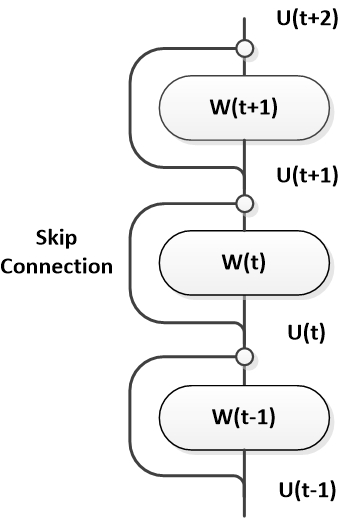}}
		\subfigure[]{
			\includegraphics[width=1.8in]{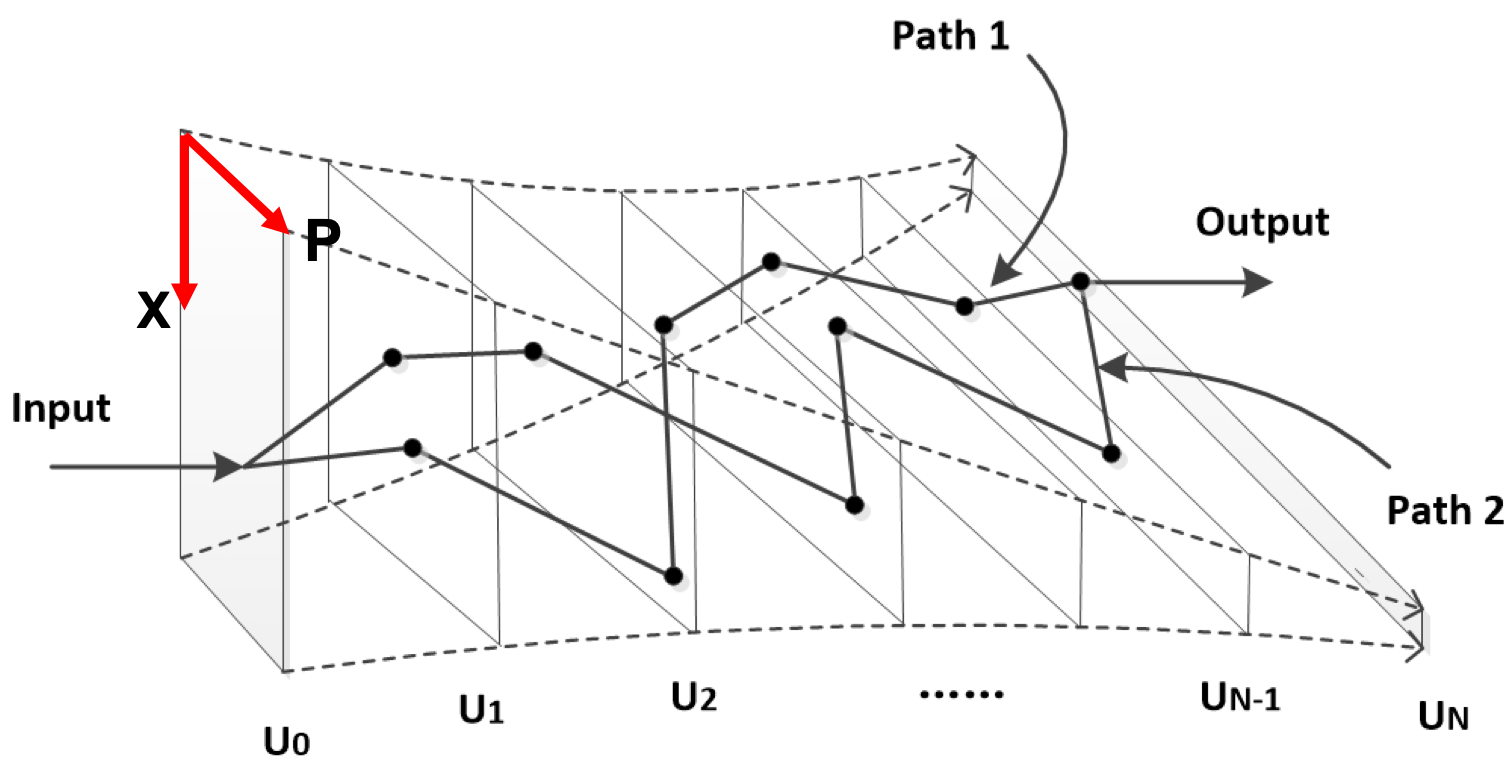}}
		\subfigure[]{
			\includegraphics[width=2.4in]{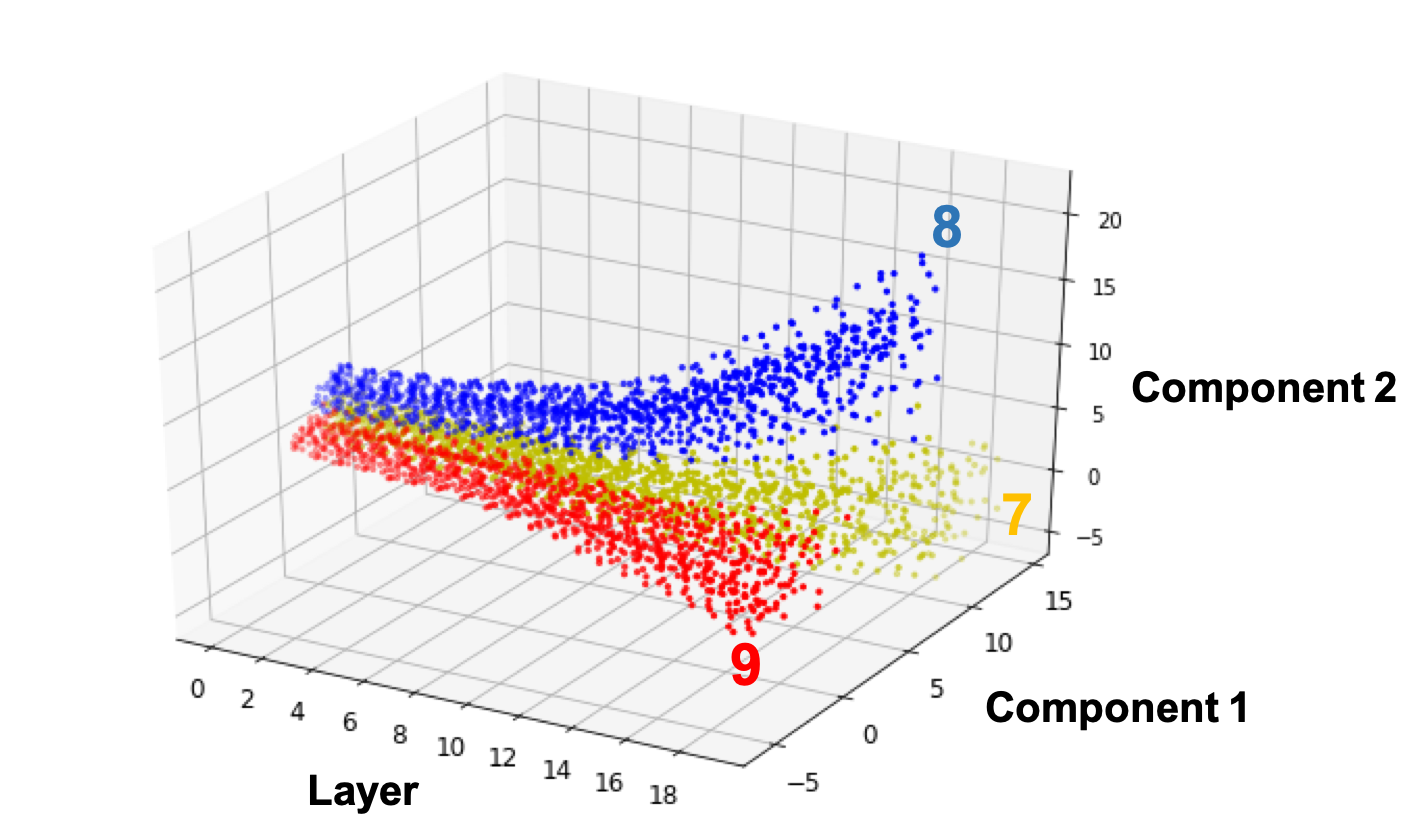}}
		\caption{(a) A path from $u(t-1)$ to $u(t+2)$ with three residual blocks. (b) A typical classification ResNet architecture, where $x$ dimension is shrinking while the $p$ dimension is expanding. There are various paths from the input to the output. (c) Different numbers from Mnist dataset moves along different trajectories in ResNet propagating process. 
	    The outputs from layers of ResNet are compressed to 2-dimension principle components with PCA.}
		\label{fig:path_and_CNN}
	\end{figure}
	\section{Comparisons Between ResNet and Traditional CNN}
	Here, we provide more analyses to demonstrate the advantages of ResNet over traditional CNN. In the training process with back propagation, the gradient descent algorithm adjusts the weights of network neurons with top-down optimization. As shown in Fig.~\ref{fig:path_and_CNN}(a), considering a path of three layer residual blocks between $u(t+2)$ and $u(t-1)$, we have
\begin{equation}
	u(t+2)=(1+w(t+1))(1+w(t))(1+w(t-1))u(t-1)
\end{equation}
	Here, we assume that the disturbance $\triangle w(t-1)$ is imposed to the $(t-1)$-layer convolution kernel, such that the corresponding distortion of $u(t+1)$ is:
	\begin{small}
		\begin{equation}
		\begin{aligned}
		&\triangle u(t+2)=\\
		&\lbrace\frac{\partial u(t+2)}{\partial w(t+1)}\frac{\partial w(t+1)}{\partial w(t)}\frac{\partial w(t)}{\partial w(t-1)}+ \frac{\partial u(t+2)}{\partial w(t+1)}\frac{\partial w(t+1)}{\partial w(t-1)} \\
		&+\frac{\partial u(t+2)}{\partial w(t)}\frac{\partial w(t)}{\partial w(t-1)}+\frac{\partial u(t+2)}{\partial w(t-1)}\rbrace \triangle w(t-1) u(t-1)
		\label{con:bp}
		\end{aligned}
		\end{equation}
	\end{small}
	Regarding the back propagation process, $\triangle u(t+2)\ / \triangle w(t-1)$ is calculated to adjust the weight of $w(t-1)$. As such, there is a skip connection between $w(t-1)$ and $u(t+2)$ which corresponds to the term $\frac{\partial u(t+2)}{\partial w(t-1)}$ in Eqn.~(\ref{con:bp}). As network goes deeper, other terms of back propagation chain tend to approach zero except for this skip connection term. 
	This explains how ResNet deals with the gradient vanishing issue. On the contrary, without skip connection, the convolution kernel can be approximated with $g(t)=e^{w(t)}$, and considering the path between $u(t+2)$ and $u(t-1)$:
	\begin{small}
		\begin{equation}
		\begin{aligned}
		&u(t+2)=g(t+1)g(t)g(t-1)u(t-1) \\
		&\triangle u(t+2)= \frac{\partial u(t+2)}{\partial g(t+1)} \frac{\partial g(t+1)}{\partial g(t)} \frac{\partial g(t)}{\partial g(t-1)} \triangle g(t-1) u(t-1)
		\end{aligned}
		\end{equation}
	\end{small}
	The partial differential chains of back propagation tend to approach zero as network goes deeper, and the back propagation gradient becomes vanishing when network goes deeper. This provides the explanations why skip connections can successfully address the gradient vanishing problem in the training process.
	\section{Experiments and Analyses} 
	\subsection{Variance analysis of Resnet}
	\begin{figure}[]
		\centering
		\includegraphics[width=3.3in]{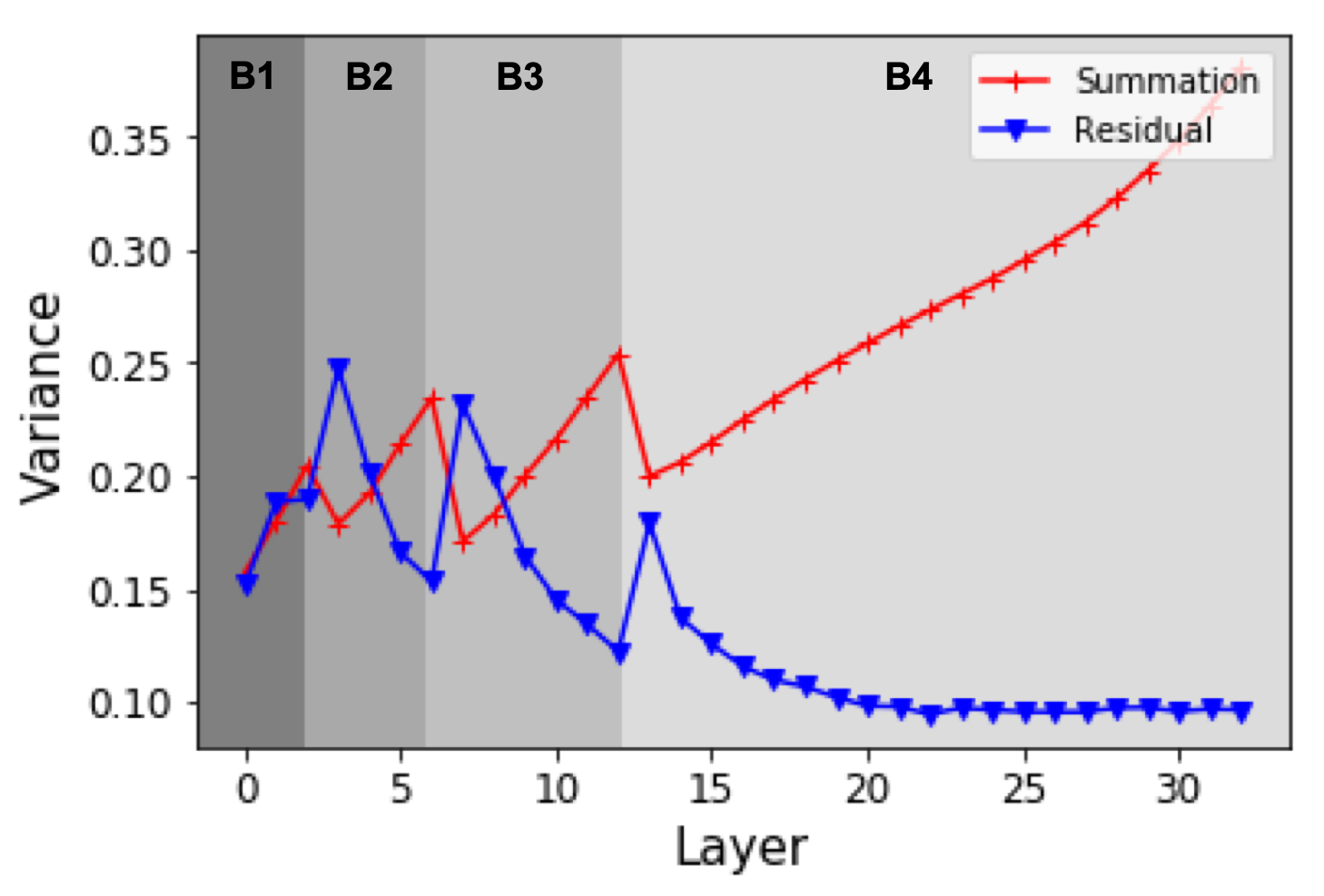}
		\caption{Illustrations of the variance of layers summing up residual and skip connection in Resnet-34 (red curve), as well as the variance of residual layers (blue curve). B1,B2,B3 and B4 represent four blocks of Resnet with different channel sizes and feature sizes.}
		\label{fig:layer_var}
	\end{figure}
	We have illustrated that the convolution process of the residual block is equivalent to extracting the partial differential part of PDE, and demonstrated that the residual network can be formulated based on the path integral form. Based on these derivations, the variance of each feature map after convolution is further explored. According to the initialization in \cite{he2015delving}, the weights of convolution are initialized using a scaling factor of $\sqrt{2/N}$, where \(N\) denotes the layer dimension of the input layer for the convolution kernels. The weights of the kernels follow Gaussian distribution with zero bias. Such initialization aims to ensure that the variance of the input layer equals to the variance of the output layer. However, the experimental results of the residual network does not accord with the equal variance assumption, as shown in Fig.~\ref{fig:layer_var}. In particular, the variances of output layers grow as the network goes deeper. However, variances of layers after convolution naturally decrease in the network. The decreasing trend of the blue curve originates from the natural property of partial differential equation stated in section 3, as the variance of a stochastic diffusion process tends to decrease as its entropy increases. Let $t$ represent the depth of residual layer, and the variance of layers after convolution is approximately proportional to $1/t$. According to the path integral formula derived from section 4,  the variance of final output layers is proportional to the exponential of path integral results, which implies it is proportional to $exp(\int dt \frac{1}{t})=t$, conforming to the red curve shown in Fig.~\ref{fig:layer_var}. As such, the experimental results in Fig.~\ref{fig:layer_var} provide useful evidence that supports our theoretical analysis on Resnet.
	\subsection{PDE residual network}
	\begin{figure}[]
		\centering
		\includegraphics[width=2.5in]{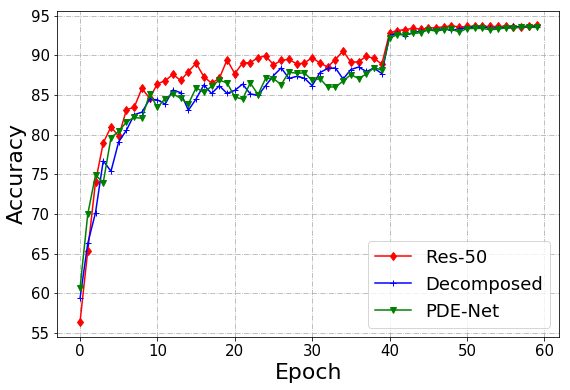}
		\caption{Classification accuracy on test dataset in the training process for ResNet, PDE-Net and Decomposed-Net. }
		\label{fig:net_acc}
	\end{figure}
	In previous sections, we have established the relationship between PDE and the convolutional residual block. In particular, we demonstrate that any two order PDE can be rewritten  as a residual block with the corresponding convolutional size equaling to $3 \times 3$. In this section, we designed two kinds of networks to validate our theory: PDE network and decomposed network. \\
	For PDE network, we replace the residual block of Resnet with sets of partial differential operators. We define kernel $[1,-2,1]$ as secondary derivative operator $\frac{\partial^{2}}{\partial x}$, define $[1,0,-1]$ as operator of derivative $\frac{\partial}{\partial x}$, and kernel $[0,1,0]$ represents the constant term. Then we initialize PDE parameters $\sigma$, $b$ and $c$ corresponding to the derivative operator above for two spatial dimensions. These three parameters are trainable. Then we design $M$ sets of PDE equations as follows,
	\begin{small}
		\begin{equation}
		\sum_{M}  \sigma^{2}_{1}\frac{\partial^{2}}{\partial x} +\sigma^{2}_{2}\frac{\partial^{2}}{\partial y} +b_{1}b_{2}\frac{\partial^{2}}{\partial x \partial y} + b_{1}c_{2}\frac{\partial}{\partial x} + b_{2}c_{1}\frac{\partial}{\partial y}  + c_{1}c_{2}
		\label{con:Mpdes}
		\end{equation}
	\end{small}
	Here, the parameters $\sigma_{1}, \sigma_{2}, b_{1}, b_{2}, c_{1}, c_{2}$ are trainable. $M$ sets of partial differential operators replace the role of convolutional residual block. For simplicity, $M$ is chosen to be equal to the number of original convolutional kernels. \\
	For decomposed network, every $3\times 3$ convolutional kernel of Resnet is decomposed into a $1 \times 3$ kernel and a $3\times 1$ kernel. In PDE network we notice that not all size 3 convolutional kernels are equivalent to a two order PDE. Because $3\times 3$ convolutional kernel has 9 independent parameters while PDE operator has only 6 parameters. From the view of independent parameters, using a set of PDE operators is equivalent to successively applying a $1\times3$ convolution and a $3\times1$ convolution, and the total number of independent parameters is 6, equalling to the two order PDE operator. It provides the idea of kernel decomposition. Therefore we designed decomposed network to observe whether cutting down parameters influence the network performance. \\
	The experimental results are shown in Table \ref{tab:acc1}. Testing accuracy during training process for three different networks are shown Fig.~\ref{fig:net_acc}. Experiments are conducted on Cifar-10 to classify images, and we adopt similar architecture from the original ResNet for PDE-net and decomposed-net. In particular, four different depth of networks are trained on Cifar-10 dataset.
	\begin{table}
		\caption{Illustrations of classification accuracy on Cifar-10.}
		\centering
		\label{tab:acc1}
		\setlength{\tabcolsep}{1mm}{
			\begin{tabular}{cccccc}
				\toprule
				Depth &ResNet&PDE-Net&Decomposed\\
				\midrule
				101& 94.89&93.38&94.68 \\
				50& 94.23 & 93.21 & 94.02 \\
				34& 93.25 & 92.97 & 93.33 \\
				18&  93.49&92.84&94.10\\
				\bottomrule
		\end{tabular}}
	\end{table}
	We find that the decomposition process of convolurional kernels can cause negligible accuracy decrease when compressing the residual network with high compression rate, and the PDE-net could also hit high classification accuracy. The experiment fully demonstrates the effectiveness of our theory analysis, as replacing all $3\times3$ convolutional kernels with successive $1\times3$ kernels and $3\times1$ kernels does not lead to classification accuracy degradation. Furthermore, PDE-nets replacing convolutional residual block with partial differential operators can successfully achieve similar classification accuracy comparing to Resnets.
	\subsection{Robustness analyses of residual block}
	Here, based on our theoretical analyses, validations are further provided to demonstrate the advantages of ResNet over traditional CNN. As stated in Eqn.~(\ref{con:FPDES}), the residual block can be rewritten as an exponential function in the frequency domain, because the weights of convolution kernels are much smaller than weights of shortcut connections in the residual block. As for ordinary convolution without skip connection, there does not exist any exponential function and the forward propagation could be expressed as $\tilde{u}(p,t)=\hat{H}_{p}t \cdot \tilde{u}(p,0)$ in the frequency domain. Here $\hat{H}_{p}$ stands for the ordinary convolution kernel without skip connection. In the training and testing process, since the convolutional kernels could be subject to certain perturbations, the robustness of ResNet and ordinary CNN without skip connections is further compared in this scenario. In particular, we assume that the noise on convolutional kernels are proportional to the absolute values of kernel weights, which implies that $\delta T \sim \hat{T}_{p}t$ and $\delta H \sim \hat{H}_{p}t$. Here, $\delta T$ and $\delta H$ stand for the perturbation on weights of residual convolution kernel and ordinary convolution kernel. Since $\hat{T}_{p}t \ll 1$, we have $\delta T \ll \exp(\hat{T}_{p}t)$. However, for ordinary convolution CNN, $\delta H$ is proportional to the propagator $\hat{H}_{p}t$, which means that noise on ordinary convolutional kernels could have more significant influence comparing to residual blocks.
	
	\begin{figure}
		\centering
		\includegraphics[width=2.5in]{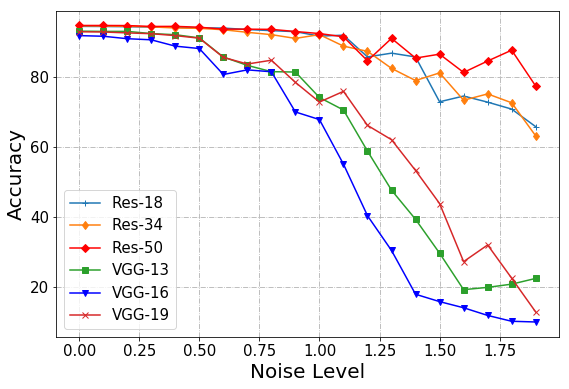}
		\caption{Test accuracy of ResNet and VGG network when the noise is proportionally injected to the convolution filter weights.}
		\label{fig:robust}
	\end{figure}
	
	From Fig.~\ref{fig:robust}, we can see that ResNet could be much more robust than ordinary convolutional neural networks. When certain noise proportional to the absolute value of convolutional weights are added, the accuracy of VGG-16 decreases much faster than that of ResNet-18. The experimental results are in accordance with our theoretical analyses on the advantage of the residual block.

	\section{Conclusion}
	Several important mathematical properties of ResNet have been investigated in this paper. It is shown that a residual block can be formulated by a certain PDE. With this conclusion, the ResNet is demonstrated to be equivalent to path integral formulation, which provides useful evidence regarding the superiority of ResNet over ordinary CNNs from the perspective of gradient vanishing. These properties revealed here may find wide applications in the CNN design and expedite the explorations of the emerging research areas. 
	\nocite{langley00}
	
	\bibliography{example_paper}
	\bibliographystyle{icml2019}


\end{document}